\documentclass{article}

\PassOptionsToPackage{table,dvipsnames}{xcolor}
\usepackage{colm2024_conference}

\usepackage{graphicx}
\usepackage{float}
\usepackage{placeins}
\usepackage{setspace}
\usepackage{titling}
\usepackage{natbib}
\usepackage{amsmath}
\usepackage{amssymb}
\usepackage{algorithm}
\usepackage{comment}
\usepackage{algpseudocodex}
\usepackage{multirow}
\usepackage{booktabs}
\usepackage{array}
\usepackage{makecell}
\usepackage{subcaption}
\usepackage{fancybox}
\usepackage{fancyvrb}
\usepackage{hyperref}
\tcbuselibrary{skins,breakable}

\usepackage{url}            
\usepackage{booktabs}       
\usepackage{multirow}    
\usepackage{amsfonts}       
\usepackage{nicefrac}       
\usepackage{microtype}      
\usepackage{natbib}
\usepackage{enumerate}
\usepackage{hhline}
\usepackage{makecell}
\usepackage{pifont}

\usepackage{graphicx} 
\usepackage{amsmath}
\usepackage{amsthm}
\usepackage{amssymb}
\usepackage{tikz}
\usepackage{xcolor}
\usetikzlibrary{arrows}

\allowdisplaybreaks

\usepackage{mathrsfs}

\usepackage{hyperref}
\usepackage{bm}

\allowdisplaybreaks

\newtheorem{principle}{Principle}

\newtheorem{defn}{Definition}

\usepackage[capitalize,noabbrev]{cleveref}
\crefname{thm}{Theorem}{Theorems}
\crefname{lem}{Lemma}{Lemmas}
\crefname{cor}{Corollary}{Corollaries}
\crefname{prop}{Proposition}{Propositions}
\crefname{asmp}{Assumption}{Assumptions}
\crefname{defn}{Definition}{Definitions}
\crefname{oracle}{Oracle}{Oracles}
\crefname{fact}{Fact}{Facts}
\crefname{conj}{Conjecture}{Conjectures}
\crefname{rem}{Remark}{Remarks}
\crefname{example}{Example}{Examples}
\crefname{condition}{Condition}{Conditions}
\crefname{exercise}{Exercise}{Exercises}
\crefname{algorithm}{Algorithm}{Algorithms}
\crefname{table}{Table}{Tables}
\crefname{figure}{Figure}{Figures}
\crefname{section}{Section}{Sections}
\crefname{subsection}{Section}{Sections}
\crefname{appendix}{Appendix}{Appendices}
\crefname{message}{Message}{Messages}

\definecolor{red}{rgb}{1, 0, 0}

\definecolor{green}{rgb}{0, 1, 0}

\definecolor{blue}{rgb}{0, 0, 1}

\definecolor{orange}{rgb}{1, 0.4, 0.0}

\input{packages/math_commands}

\usepackage{listings}
\hypersetup{
  colorlinks   = true, 
  urlcolor     = blue, 
  linkcolor    = blue, 
  citecolor   = blue 
}

\definecolor{lightgray}{gray}{0.95} 

\title{When Do Larger Batches Help Scale LLM\\Reinforcement Learning?}

\author{%
  \bfseries\large
  Ziniu Li \quad Jinbo Wang \quad Guanhua Huang \\
  Feiyuan Zhang \quad Pengbo Li \quad Alex Chen \\ 
  \vspace{0.4cm}
   Tencent Hunyuan Team \\
   \vspace{0.15cm}
   {\normalfont\normalsize\texttt{ziniuli@tencent.com}}
}

\begin{document}
\maketitle

\begin{abstract}
Larger batches reduce the variance of stochastic gradients per update and are therefore often expected to accelerate training. Yet whether this statistical benefit translates into lower wall-clock time-to-target remains unclear, because each update consumes more samples and may take longer to execute. We study this tradeoff in reinforcement learning for large language models. We separate its \emph{algorithmic} and \emph{systems} effects by comparing learning and execution along their natural axes. At the algorithmic level, we compare configurations at equal \emph{cumulative} sample counts while retuning batch-dependent hyperparameters. Over a bounded range of batch sizes, this procedure yields an approximately \emph{batch-size-invariant} family whose members follow similar sample-indexed learning trajectories. At the systems level, we exploit the computational \emph{asymmetry} between rollout generation and training: autoregressive generation is often memory-bandwidth-bound at low concurrency, whereas training work scales approximately with the number of processed tokens. Combining these two views yields a direct decision rule: a larger-batch configuration reduces time-to-target only when its throughput gain exceeds its samples-to-target penalty. Experiments with GRPO and PPO support both sides of this decomposition. At the algorithmic level, square-root learning-rate scaling with Adam produces approximately batch-size-invariant learning curves over a bounded range of batch sizes. At the systems level, larger batches improve generation throughput by up to \(2.29\times\) on fixed hardware. In GRPO, combining higher throughput with learning-rate retuning reduces time-to-target by up to 29\%, whereas increasing the batch without retuning is slower despite its higher throughput.
\end{abstract}

\begin{figure}[H]
    \centering
    \includegraphics[width=0.94\textwidth]{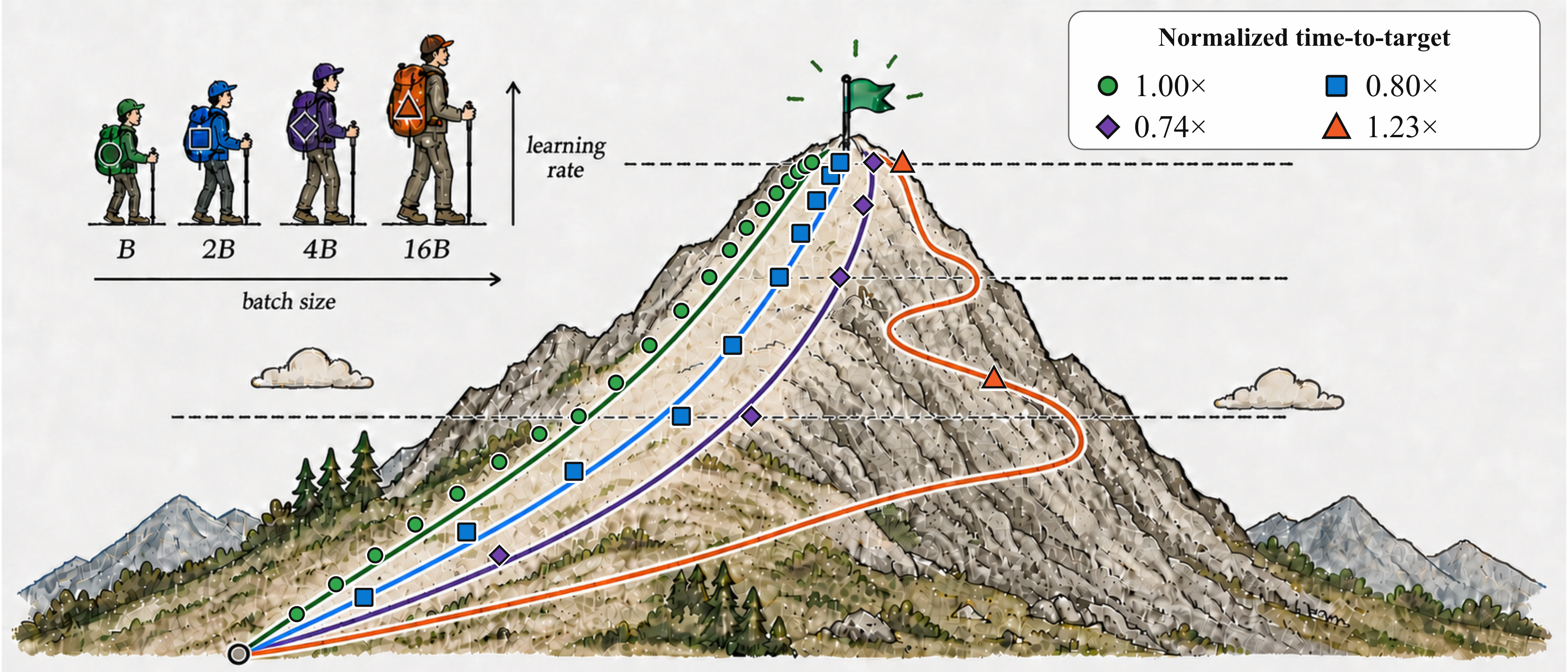}
    \caption{A data-informed schematic based on selected GRPO operating points from our experiments (Table~\ref{tab:grpo-target-accounting}). Each trail marker denotes one optimizer update. It highlights three messages: (1) after learning-rate retuning, $B$, $2B$, and $4B$ are approximately batch-size invariant in cumulative samples; (2) higher realized throughput within this regime improves wall-clock efficiency; and (3) beyond this regime, as at $16B$, normalized time-to-target deteriorates, rising above the $B$ baseline.}
    \label{fig:overview-illustrated}
\end{figure}

\section{Introduction}
\label{sec:introduction}

Scaling reinforcement learning (RL) to larger language models (LLMs) and longer training runs demands substantially more computation. Recent reasoning systems such as OpenAI o1 and DeepSeek-R1 illustrate the growing role of RL post-training in improving LLM reasoning \citep{openai2024o1, deepseekai2025deepseekr1}. Increasing the batch size is a primary means of exposing parallelism: larger batches place more rollout-generation and optimization work in flight, potentially improving hardware utilization while reducing stochastic-gradient variance \citep{mccandlish2018empirical,shallue2019measuring}. However, each update also consumes more newly generated samples and may take longer to collect and optimize. This raises a fundamental question: can large-batch training actually reduce wall-clock time-to-target?

Prior studies \citep{goyal2017accurate,smith2018dont,shallue2019measuring} suggest that data-parallel scaling is most effective when two conditions hold: (1) additional computational resources allow larger batches to be processed through data parallelism while keeping per-step time approximately constant; and (2) the batch remains within the statistically efficient regime, typically below or near a threshold commonly referred to as the \emph{critical training batch size}, with batch-dependent hyperparameters appropriately retuned. In this paper, we instead study this question in the practically important \emph{fixed-hardware} setting\footnote{Focusing on fixed hardware isolates the effect of batch size from that of adding accelerators. The same throughputversus-sample-efficiency criterion also applies when the accelerator count changes, after accounting for the corresponding change in throughput.}: \textbf{can increasing the batch size alone reduce wall-clock time-to-target, and if so, under what conditions?} In conventional supervised learning, this opportunity is often limited once dense forward–backward computation saturates the available hardware. LLM RL has a different computational structure: training samples must first be generated, typically through autoregressive decoding, before they can be consumed by model updates \citep{li2024remax,shao2024deepseekmath}. Consequently, batch tuning couples an \emph{algorithmic} question (how much learning is obtained from a fixed number of samples) with a \emph{systems} question (how quickly those samples are generated and consumed). 

To formalize the tradeoff, we use a two-level learning-effectiveness–throughput decomposition:
\begin{equation}
    \underbrace{\frac{\mathrm{d}J}{\mathrm{d}t}}_{\text{learning progress rate}}
    =
    \underbrace{\frac{\mathrm{d}J}{\mathrm{d}N}}_{\text{sample efficiency}}
    \times
    \underbrace{\frac{\mathrm{d}N}{\mathrm{d}t}}_{\text{systems efficiency}},
    \label{eq:learning-throughput-decomposition}
\end{equation}
where $J$ denotes evaluation performance, $N$ denotes cumulative consumed samples, and $t$ denotes wall-clock time. The first factor measures learning progress per sample and captures the algorithmic effect of batch size; the second measures the rate at which samples are generated and consumed and captures its systems effect. To compare batch configurations, we fix a performance target. For each configuration, let the \emph{samples-to-target} $N^\star$ be the cumulative number of samples required to reach it, and let $T$ be the corresponding wall-clock time-to-target. The realized average end-to-end throughput over that interval is then $q=N^\star/T$. Thus, comparing a larger-batch configuration $2$ with a reference configuration $1$,
\begin{equation}
    T_2<T_1
    \quad\Longleftrightarrow\quad
    \frac{q_2}{q_1}
    >
    \underbrace{\frac{N_2^\star}{N_1^\star}}_{r_N},
    \label{eq:intro-throughput-sample-penalty}
\end{equation}
where $r_N$ is the samples-to-target penalty. 

Equation~\eqref{eq:intro-throughput-sample-penalty} answers the title question: a larger batch helps precisely when its throughput gain exceeds its sample penalty. In practice, we can implement this decision rule in two stages. First, we retune batch-dependent hyperparameters to construct an approximately \emph{batch-size-invariant} family, for which $r_N\approx1$ over a bounded regime \citep{mccandlish2018empirical,hilton2022batch,malladi2022sde}. Second, we exploit generation--training asymmetry to increase $q$. In bandwidth-bound decoding, model weights are streamed once per step, so larger active batches amortize this nearly fixed traffic across more tokens. Generation time can therefore grow \emph{sublinearly} with batch size while training work grows approximately linearly with processed tokens \citep{williams2009roofline,pope2023efficiently,zhong2024distserve}. This asymmetry can increase end-to-end throughput.

Empirically, we study GRPO with Qwen3-30B-A3B-Instruct-2507 \citep{yang2025qwen3technicalreport} and PPO with an internal Hunyuan mixture-of-experts model with 3B active parameters. Figure~\ref{fig:overview-illustrated} provides a schematic preview.  First, square-root learning-rate scaling for Adam produces approximately batch-size-invariant learning curves in both workloads. In particular, this invariance holds only over a \emph{bounded} range and begins to break down at extremely larger batch sizes. Further, this alignment also depends on batch-dependent learning-rate retuning:
in our GRPO fixed-learning-rate control, doubling the batch increases samples-to-target by 67\%. Second, we show that increasing the PPO batch by $4\times$ on fixed hardware improves generation throughput by $2.29\times$. In GRPO, larger batches with learning-rate retuning reduce time-to-target by up to 29\%. Together, these results demonstrate that fixed-hardware speedup is conditional
on both preserving learning per sample and improving systems throughput.

To summarize, our contributions are:
\begin{itemize}
    \item We formulate batch-size invariance as a comparison and transfer principle and study it empirically across PPO and GRPO in modern LLM RL post-training.
    \item We characterize the generation--training asymmetry that enables acceleration even under a fixed hardware allocation and measure how rollout-collection cost changes with batch size.
    \item We operationalize the decision rule as a practical two-stage tuning procedure. Our experiments support both components and both directions of the rule: generation throughput improves by up to $2.29\times$, tuned larger-batch configurations reduce time-to-target by up to 29\%, and a fixed-learning-rate control is slower despite higher throughput.
\end{itemize}

\section{Problem Formulation}
\label{sec:formulation}

The criterion in Equation~\eqref{eq:intro-throughput-sample-penalty} combines a samples-to-target ratio with a throughput ratio. We therefore begin by fixing the sample-accounting rule and distinguishing the batch quantities that affect learning from those that affect execution. At a rollout-collection boundary, let $\theta$ denote the current policy parameters. The policy $\pi_{\theta}$ samples $G$ responses for each of $P$ prompts,
\begin{equation*}
    y_{i,j} \sim \pi_{\theta}(\cdot \mid x_i),
    \qquad i \in \{1,\ldots,P\},\quad j \in \{1,\ldots,G\}.
\end{equation*}
After a reward model or verifier scores these rollouts, the algorithm constructs returns or advantage estimates. PPO trains an actor together with a learned value model \citep{schulman2017ppo}. Value-model-free, REINFORCE-style methods such as ReMax instead avoid training a separate critic and construct a variance-reduction baseline from sampled rewards \citep{williams92,li2024remax}, while GRPO uses responses to the same prompt to form a group-relative baseline \citep{shao2024deepseekmath}.

This notation covers one-response objectives such as PPO and ReMax with $G=1$, as well as grouped objectives such as GRPO with $G>1$. One optimizer update uses a nominal training batch $B_{\mathrm{train}}=PG$. The generation-side batch $B_{\mathrm{gen}}$ equals $PG$ in a coupled pipeline but may be larger under asynchronous streaming. Micro-batching and gradient accumulation affect execution only.

We use $N$ for cumulative training samples. For an \emph{algorithmic configuration}---the training batch and its batch-dependent hyperparameters, including the learning rate---the sample-indexed curve $J(N)$ defines samples-to-target $N^\star$. A \emph{systems realization}---generation concurrency, placement, runtime, and hardware---determines the realized end-to-end throughput $q=N^\star/T$, and hence $T=N^\star/q$. We use $q_{\mathrm{gen}}$ specifically for generation-stage throughput.

\section{Training-Side Comparability: Invariance and the Critical Batch}
\label{sec:training-scaling}

Recall the framework we follow in this paper: Equation~\eqref{eq:learning-throughput-decomposition} decomposes wall-clock learning progress into learning per sample and systems throughput. Determining whether a larger batch improves wall-clock progress therefore requires separating its training-side sample efficiency from its systems-side throughput. In particular, comparisons at equal optimizer updates are inappropriate because different batch sizes correspond to different sample budgets: after \(k\) updates, a run with batch size \(2B\) has processed \(2kB\) samples, whereas a run with batch size \(B\) has processed only \(kB\). Better performance in the larger-batch run may therefore reflect greater sample exposure rather than higher sample efficiency.

These considerations motivate \emph{batch-size invariance}: after admissible retuning, learning trajectories should remain approximately aligned when indexed by cumulative samples, even though larger batches require fewer sequential updates. In this regime, samples-to-target are approximately preserved, so $r_N\approx1$, and any wall-clock gain can be attributed to improved throughput. We next formalize this condition and describe how we construct and test it.

\subsection{Batch-Size Invariance and Transfer}
\label{sec:batch-size-invariance}

Throughout this section, $B$ denotes the training batch. After applying a batch-dependent retuning rule, let $J_B(N)$ denote the expected evaluation performance after $N$ cumulative training samples. Batch-size invariance asks whether the hyperparameters can be retuned as $B$ changes so that these sample-indexed trajectories remain approximately aligned \citep{hilton2022batch}.

\begin{defn}[Approximate batch-size invariance]
\label{def:batch-size-invariance}
Fix a model, task, algorithm, reward definition, evaluation protocol, reference batch $B_0$, an admissible batch-retuning rule, sample interval $\mathcal N$, and tolerance $\varepsilon>0$. The resulting batch-dependent family is approximately invariant if, for every batch $B$ in the family,
\begin{equation*}
    \sup_{N\in\mathcal N}
    \left|
    J_B(N)-J_{B_0}(N)
    \right|
    \leq\varepsilon.
\end{equation*}
The expectation over rollout, optimization, and evaluation randomness is implicit in $J_B$.
\end{defn}

Performance-level invariance allows distinct optimization paths and learned policies as long as they attain similar expected evaluation performance. Related work likewise distinguishes agreement in loss from agreement in function space when studying batch-size effects in online learning \citep{vyas2024beyond}. Note that the concept of batch-size invariance is deliberately local: changing the model, data distribution, objective, reward, or training stage defines a new learning problem and requires re-establishing alignment.

Why is this property important in practice? It enables data-parallel speedup without sacrificing learning per sample: under a fixed sample budget, increasing the batch size by a factor of \(s\) reduces the number of sequential optimizer updates by the same factor $s$. For example, \citet{goyal2017accurate} reduced ResNet-50/ImageNet training from 29 hours with a batch of 256 on 8 GPUs to 1 hour with a batch of 8192 on 256 GPUs, while matching accuracy; at the fixed 90-epoch sample budget, the $32\times$ larger batch required $32\times$ fewer sequential optimizer steps. Taking this idea further, the critical-batch literature asks how far this reduction in sequential optimizer steps can be sustained as batch size increases before diminishing returns set in \citep{mccandlish2018empirical,shallue2019measuring}.

\subsection{From Invariance to the Critical Training Batch}
\label{sec:critical-training-batch}

Batch-size invariance describes the ideal pre-knee regime: if samples-to-target are preserved, a larger batch converts the same sample budget into fewer sequential updates. This gain \emph{cannot} continue indefinitely. Under independent sampling, gradient averaging ideally reduces stochastic-gradient variance as $1/B$. This benefit diminishes once the reducible noise is small relative to the gradient signal, or when correlations and nonstationarity leave residual variation. Further averaging then yields little improvement in update quality. The critical training batch marks this onset of diminishing returns for a declared learning target \citep{mccandlish2018empirical,shallue2019measuring}.

\begin{defn}[Critical training batch]
\label{def:critical-training-batch}
Fix a model, task, algorithm, evaluation target $J^\star$, an admissible hyperparameter-retuning protocol, and a nonempty set of admissible training batches $\mathcal B$. Let $K^\star(B)$ be the minimum expected number of sequential actor updates required for the policy to reach $J^\star$ with batch $B$ under that protocol, so $1/K^\star(B)$ measures update efficiency. Define
\begin{equation*}
    B_{\mathrm{train}}^{\mathrm{crit}}
    =\min\left\{
        B\in\mathcal B:
        \frac{1}{K^\star(B)}
        \geq
        \frac{1}{2}
        \sup_{B'\in\mathcal B}\frac{1}{K^\star(B')}
    \right\}.
\end{equation*}
\end{defn}

For a fixed batch, the sample and update counts satisfy $N^\star(B)\approx B K^\star(B)$. Within an invariant family, $N^\star(B)$ is approximately preserved, and hence
\begin{equation}
    K^\star(B)\approx \frac{N^\star}{B}.
    \label{eq:invariant-update-scaling}
\end{equation}
Equation~\eqref{eq:invariant-update-scaling} predicts that doubling the batch should roughly halve the sequential updates needed to reach the target.

Under the empirical model $K^\star(B)=K_{\min}^\star(1+B_{\mathrm{train}}^{\mathrm{crit}}/B)$ of \citet{mccandlish2018empirical}, $K_{\min}^\star$ is the asymptotic minimum updates-to-target. At $B=B_{\mathrm{train}}^{\mathrm{crit}}$, $K^\star(B)=2K_{\min}^\star$, so the definition recovers the conventional half-efficiency turning point when $\mathcal B$ spans the large-batch limit.

To instantiate the retuning rule in our experiments, we vary only the learning rate with batch size. For Adam, we initialize the batch-dependent learning rate with the square-root rule motivated by the SDE analysis of \citet{malladi2022sde}:
\begin{equation*}
    \eta(B)=\eta_0\sqrt{\frac{B}{B_0}}.
\end{equation*}

The square-root rule is only an initialization, not a guarantee of invariance. Although prior work suggests that other optimizer- and algorithm-specific hyperparameters may also benefit from batch-dependent adjustment \citep{goyal2017accurate,smith2018dont,malladi2022sde}, we hold them fixed for simplicity and do not explore broader retuning here. Our test therefore asks whether learning-rate scaling alone is sufficient to align the sample-indexed curves $J_B(N)$ and recover the predicted $1/B$ scaling of $K^\star(B)$. A failure of alignment may reflect either genuine statistical saturation or a limitation of this restricted protocol. In the following sections, Section~\ref{sec:training-evidence} reports these training-side tests, while Section~\ref{sec:generation-scaling} separately evaluates the generation-side throughput gains attainable on fixed hardware.

\subsection{Empirical Evidence}
\label{sec:training-evidence}

Related batch-scaling regularities appear in the empirical large-batch model of \citet{mccandlish2018empirical} and in batch-size-invariant policy optimization by \citet{hilton2022batch}. We test whether approximate invariance also appears in LLM RL. All batch-size comparisons within each workload use a fixed hardware allocation: the machine configuration and accelerator count are held constant as the batch varies. Relative to the minibatch-based PPO/PPG setting of \citet{hilton2022batch}, we study single-pass policy-gradient methods: in both GRPO and PPO, each rollout batch supports one global optimizer step, without minibatch subdivision or rollout reuse.\footnote{Under synchronized execution with zero rollout--training mismatch, this regime is on-policy.} With $G$ responses per prompt, $N_{\mathrm{train}}=kPG$ after $k$ updates; this retained-response counter indexes all learning comparisons. PPO, the special case $G=1$, provides complementary evidence. In our training infrastructure, both workloads use asynchronous partial rollouts with streaming generation, allowing generation concurrency to exceed the per-update training batch.\footnote{In our experiments, $B_{\mathrm{train}}=PG$ and $B_{\mathrm{gen}}=2B_{\mathrm{train}}$.}

\paragraph{Measurement protocol.}
We compare learning across configurations at the observed validation checkpoints. The $\pm1$-percentage-point band serves only as a descriptive visual reference for assessing approximate alignment at these checkpoints. As an indirect proxy for gradient variance, we also report the logged gradient norm. By Jensen's inequality, stochasticity can inflate its expected value; when gradient noise dominates, we therefore expect the norm to decrease as batch size grows.

\paragraph{GRPO setup.}
We train Qwen3-30B-A3B-Instruct-2507 across all GRPO configurations. The training data are a filtered and evaluated subset derived from DAPO-MATH-17K \citep{yu2025dapo}; the resulting dataset used by all runs contains 4,853 rows after filtering out solve-all and solve-none prompts under a 16K-token rollout budget. We report mean@32 performance averaged over AIME 2024 and AIME 2025. In the main sweep, $G=8$ is fixed, so scaling the prompt batch $P$ scales the retained training batch $PG$ by the same factor. We sweep $P\in\{64,128,256,512,1024,2048,4096\}$ and initialize the learning rate from $(P_0,\eta_0)=(128,10^{-6})$ using $\eta=\eta_0\sqrt{P/P_0}$. This gives $\eta=\{1/\sqrt{2},1,\sqrt{2},2,2\sqrt{2},4,4\sqrt{2}\}\times10^{-6}$. Our priority sampler weights prompts by their estimated pass rates, favoring values near $0.5$. This increases the chance that a sampled group contains both successful and unsuccessful responses, thereby reducing rejection under outcome-based filtering---called dynamic sampling in DAPO \citep{yu2025dapo}. Accordingly, our empirical sample unit is a \emph{training response}: $N=N_{\mathrm{train}}$ counts responses retained for optimizer updates after filtering.

\begin{figure}[t]
    \centering
    \includegraphics[width=\textwidth]{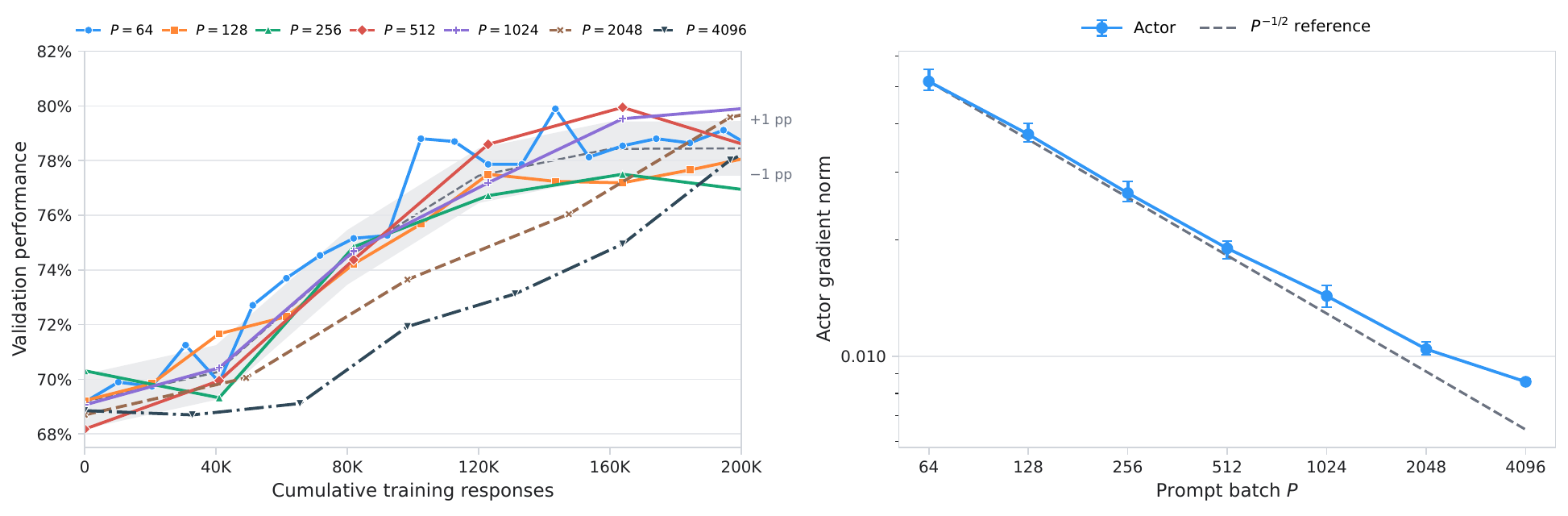}
    \caption{GRPO scaling at fixed group size $G=8$. \textbf{Left:} the gray band is a $\pm1$-percentage-point visualization band around the mean trajectory fitted at shared checkpoints spaced every 40K training responses using $P=64$--$1024$. These runs are approximately batch-size invariant over the measured window, whereas the held-out $P=2048$ and $P=4096$ trajectories deviate from the aligned family. \textbf{Right:} points and error bars show the actor-gradient-norm median and interquartile range over the common 40K--160K training-response window. The $P=64$--$1024$ medians follow the $P^{-1/2}$ reference closely, while the decline flattens at the largest batches, most clearly from $P=2048$ to $P=4096$.}
    \label{fig:grpo-training-invariance}
    \vspace{-0.35em}
\end{figure}

\paragraph{Fixed group size: invariance and its boundary.}
Figure~\ref{fig:grpo-training-invariance} (left) shows the main GRPO sweep at fixed $G=8$. After square-root learning-rate retuning, $P=64$--$1024$ follow approximately batch-size-invariant learning curves over the measured window. The $P=2048$ and $P=4096$ curves instead deviate from this aligned family, indicating that the invariant range has been exceeded.

Figure~\ref{fig:grpo-training-invariance} (right) provides a consistent optimization-side explanation. Over $P=64$--$1024$, the actor-gradient-norm proxy has a fitted log--log slope of $-0.468$, close to the $P^{-1/2}$ scaling expected when stochastic noise dominates. At the upper end, this scaling begins to flatten: doubling the batch from $P=2048$ to $P=4096$ reduces the median gradient norm to $0.823\times$, rather than the $0.707\times$ predicted by $P^{-1/2}$. The largest batches therefore no longer exhibit the same noise-dominated scaling, consistent with the loss of learning-curve invariance in the left panel.

\begin{figure}[t]
    \centering
    \includegraphics[width=\textwidth]{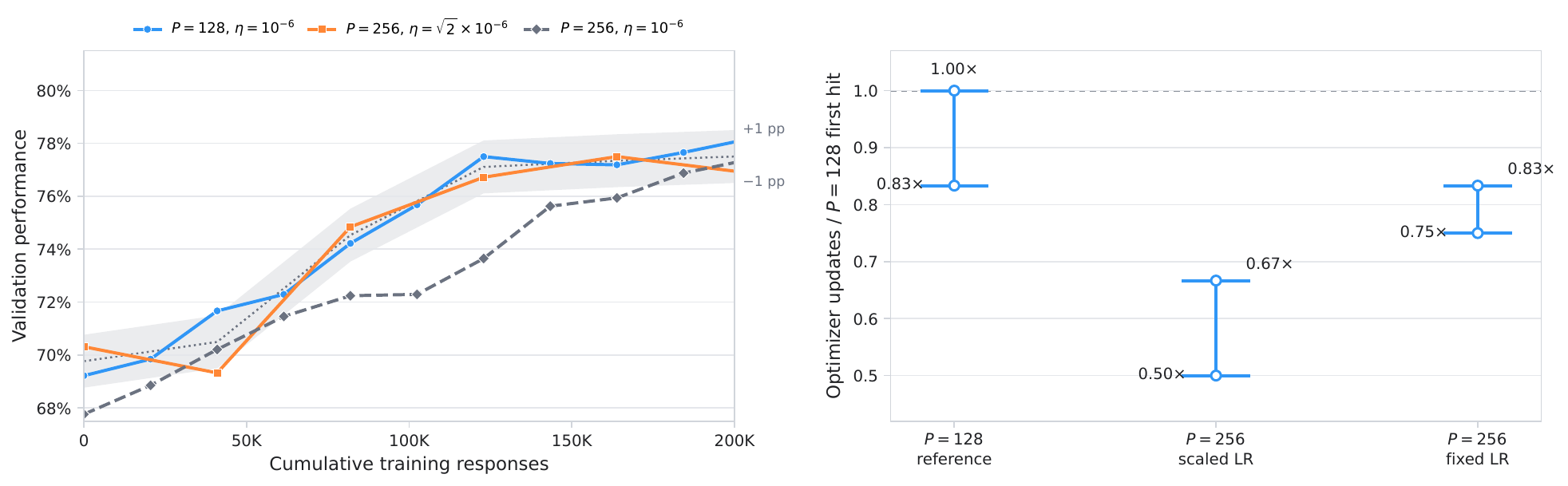}
    \caption{Learning-rate retuning improves sample-indexed alignment in GRPO. \textbf{Left:} square-root scaling keeps the $P=256$ trajectory close to the $P=128$ reference, whereas using the fixed $P=128$ learning rate slows progress. \textbf{Right:} normalized optimizer updates required to reach the target; the capped intervals show checkpoint resolution. Exact invariance predicts $0.50\times$ when the batch doubles, which is consistent with the scaled-LR run but not the fixed-LR control.}
    \label{fig:grpo-lr-ablation}
\end{figure}

\paragraph{Fixed-LR ablation: a larger batch is not automatically optimal.}
Figure~\ref{fig:grpo-lr-ablation} emulates a \emph{common but inappropriate} practice: increasing the batch while keeping the learning rate fixed. In the left panel, the scaled-LR $P=256$ run tracks the $P=128$ reference at equal cumulative training responses, whereas the fixed-LR run learns more slowly and breaks batch-size invariance. The right panel tests Equation~\eqref{eq:invariant-update-scaling}: exact invariance predicts that doubling $P$ halves the optimizer updates to target. The scaled-LR crossing interval of $0.50$--$0.67\times$ contains the ideal $0.50\times$, whereas the fixed-LR interval of $0.75$--$0.83\times$ shows a smaller gain. Thus batch scaling alone yields weaker update-side scaling and is slower than the reference in the end-to-end accounting of Section~\ref{sec:generation-evidence}.

\begin{figure}[htbp]
    \centering
    \includegraphics[width=\textwidth]{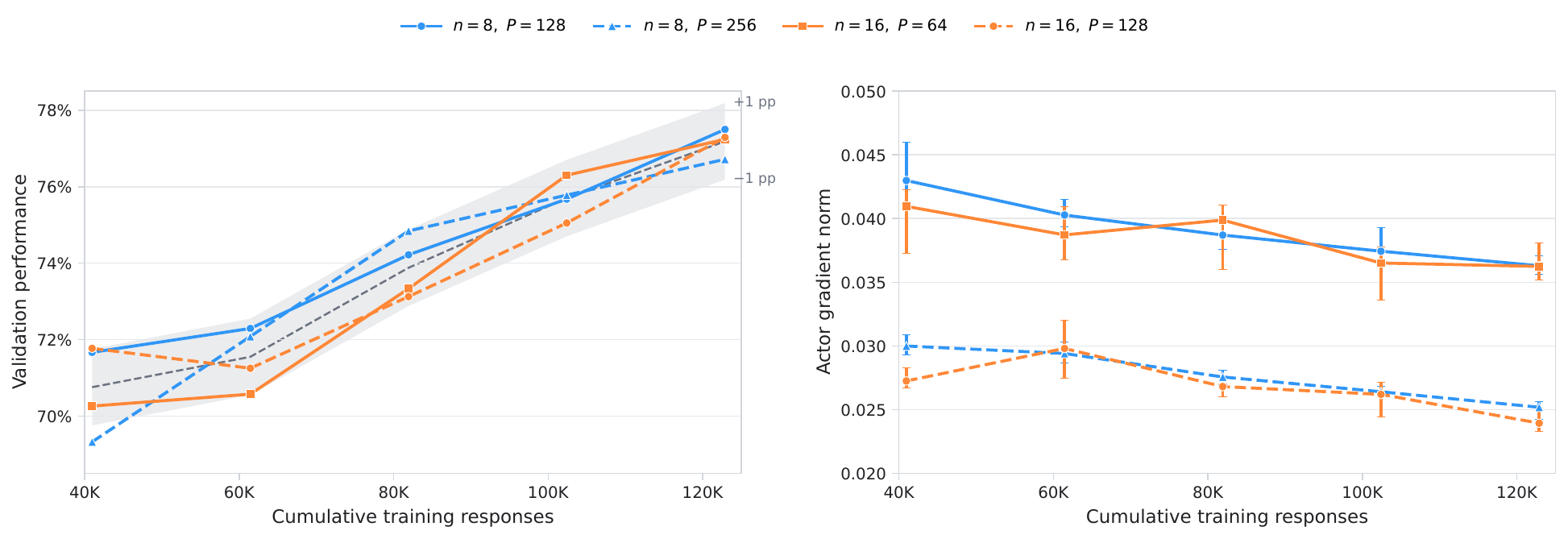}
    \caption{GRPO prompt-batch and group-size comparisons at matched training-response budgets over the common 40.96K--122.88K response window. The pairs $(G=8,P=128)$ versus $(G=16,P=64)$ and $(G=8,P=256)$ versus $(G=16,P=128)$ hold $PG$ fixed at 1024 and 2048 retained responses per update, respectively. \textbf{Left:} validation trajectories approximately align when indexed by cumulative training responses. \textbf{Right:} actor-gradient-norm summaries are also similar within each matched pair.}
    \label{fig:grpo-group-size-invariance}
\end{figure}

\paragraph{What constitutes the batch size in GRPO?}
The preceding sweep changes the prompt batch $P$ while keeping the group size $G$ fixed. A GRPO update, however, contains $P$ prompt groups with $G$ responses per group, so its batch size could refer either to the number of prompts $P$ or to the total response batch $PG$. To distinguish these possibilities, we vary $G$ while holding $PG$ fixed. Figure~\ref{fig:grpo-group-size-invariance} shows that configurations with different $(P,G)$ pairs but the same $PG$ follow similar learning curves when indexed by cumulative training responses; their gradient-norm summaries are also similar. Approximate batch-size invariance therefore continues to hold when the group size changes over the measured range. For this recipe, the relevant empirical batch and sample unit is the total number of training responses $P\times G$, rather than the prompt count $P$ alone. We test $G\in\{8,16\}$ and leave broader group sizes to future work.

\paragraph{PPO as complementary evidence.}
We next test bounded invariance in PPO, where each prompt contributes one response. We use an internal Hunyuan mixture-of-experts model and value model, each with 3B active parameters. The filtered set contains 13,405 mathematics, science, and logic examples. Starting from $(B_0,\eta_0)=(512,10^{-6})$, we sweep $B\in\{256,512,1024,2048,4096\}$ with square-root learning-rate scaling; the critic learning rate is $10\eta$. Evaluation reports held-out mean@4 on these domains. We exclude the $30\times B$ responses generated during the 30-step critic-only warm-up and re-zero the actor-training sample axis for the optimization-side comparison.

\begin{figure}[t]
    \centering
    \includegraphics[width=\textwidth]{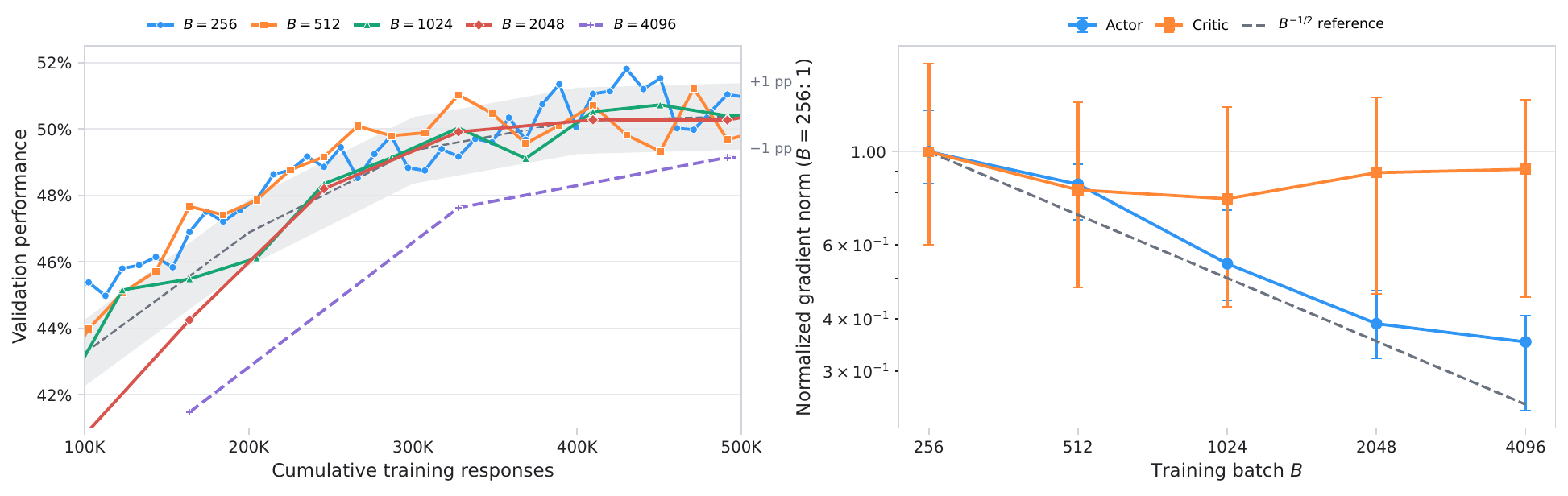}
    \caption{Training-side batch scaling for PPO. The cumulative-training-response axis excludes the $30\times B$ responses collected during each run's 30-step critic-only warm-up and is re-zeroed when actor optimization begins. \textbf{Left:} $B=256$--$2048$ approximately align, whereas the held-out $B=4096$ trajectory deviates. \textbf{Right:} normalized post-warm-up actor gradient norm decreases with batch, whereas the critic is comparatively batch-flat and noisier.}
    \label{fig:ppo-training-invariance}
\end{figure}

\paragraph{PPO results and the boundary of alignment.}
Figure~\ref{fig:ppo-training-invariance} shows approximate alignment for $B=256$--$2048$ and deviation at $B=4096$ (left), together with the corresponding optimization-side signal (right).
Over $B=256$--$2048$, the actor-gradient-norm proxy has a fitted log--log slope of $-0.471$, close to the $B^{-1/2}$ scaling expected in a noise-dominated regime. Doubling the batch from $B=2048$ to $B=4096$, however, reduces the median to only $0.905\times$, rather than the $0.707\times$ predicted by $B^{-1/2}$. This flattening is consistent with the loss of learning-curve invariance in the left panel. PPO is nevertheless more complex than GRPO because it jointly optimizes an actor and a value model, whose batch scales need not coincide.

\paragraph{Actor--critic batch scales.}
Definition~\ref{def:critical-training-batch} applies to policy learning: $K^\star(B)$ counts actor updates required to reach the validation target. PPO additionally trains a value model with its own regression objective, so the actor and critic need not share the same batch scale. Surprisingly, Figure~\ref{fig:ppo-training-invariance} (right) shows a marked asymmetry: as the batch grows, the actor gradient norm decreases, whereas the critic gradient norm remains nearly flat and substantially noisier. This suggests that larger batches benefit the actor and critic differently. Determining the critic's own critical batch requires a separate analysis of value learning, which we leave to future work.

\begin{tcolorbox}[
  enhanced,
  title={Summary and takeaway},
  fonttitle=\bfseries,
  colback=RoyalBlue!4,
  colframe=RoyalBlue!55!black,
  boxrule=0.6pt,
  arc=1.5mm,
  left=2mm,
  right=2mm,
  top=1.5mm,
  bottom=1.5mm
]
\begin{itemize}
  \setlength{\topsep}{0pt}
  \setlength{\itemsep}{0.35em}
  \setlength{\parsep}{0pt}
  \item With Adam, square-root learning-rate scaling yields approximate invariance over bounded batch-size ranges in GRPO and PPO.
  \item A fixed learning rate breaks this alignment and should be retuned when the batch changes.
  \item For measured GRPO group sizes, total responses $P\times G$ are the batch and sample unit.
  \item PPO may have distinct critical training batch sizes for its actor and critic.
\end{itemize}
\end{tcolorbox}

\section{Generation--Training Asymmetry and Fixed-Hardware Efficiency}
\label{sec:generation-scaling}

Section~\ref{sec:training-scaling} identified a bounded regime in which larger batches preserve learning per response after learning-rate retuning. We now turn to the systems question: \textbf{on fixed hardware, can a larger batch convert this algorithmic invariance into a lower time-to-target?} Classical large-batch scaling relies on additional accelerators to process larger batches in parallel. Our setting instead keeps hardware fixed, so any wall-clock gain must come from higher realized throughput.

RL creates such an opportunity because every training response must first be generated autoregressively. The key systems mechanism is a \emph{generation--training asymmetry}: doubling the rollout batch doubles the number of requested responses, but often increases generation time by substantially less than $2\times$. Autoregressive decoding is commonly underfilled at low concurrency, so additional active sequences share the cost of streaming model weights and increase response throughput. Training, by contrast, processes every token, so its arithmetic grows with training volume.

This asymmetry creates an acceleration opportunity even when the learning curve per response is unchanged. Within a batch-size-invariant range, higher response throughput directly reduces time-to-target; outside that range, the systems gain must be weighed against the samples-to-target penalty. Section~\ref{sec:generation-evidence} first presents fixed-hardware generation-stage evidence. Section~\ref{sec:sublinear-generation-cost} then explains why generation time can grow sublinearly with batch and where the gain saturates.

\subsection{Empirical Evidence}
\label{sec:generation-evidence}

The systems measurements below use the PPO and GRPO workloads introduced in Section~\ref{sec:training-evidence} and a subset of the batch configurations evaluated there. Section~\ref{sec:training-evidence} compared learning at equal cumulative responses; here we hold each workload's hardware allocation fixed and measure how increasing the batch changes rollout-collection time, actor-update time, and realized response throughput.

\begin{figure}[htbp]
    \centering
    \includegraphics[width=\textwidth]{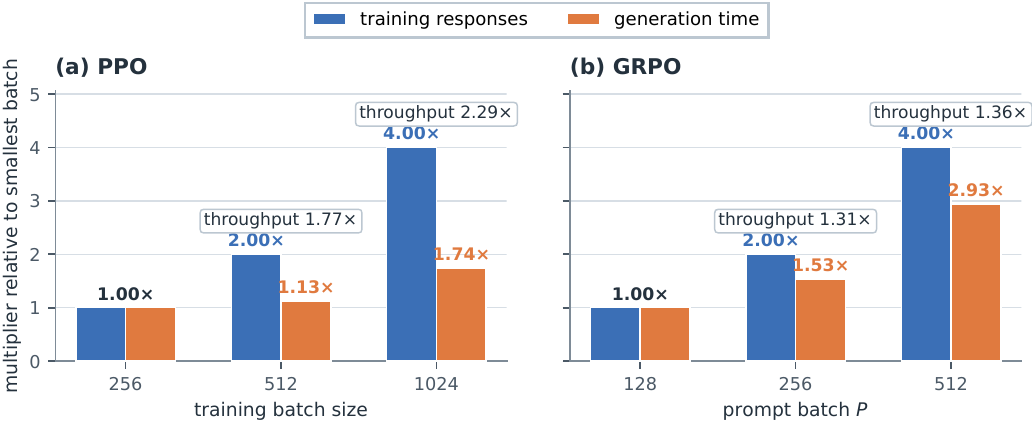}
    \caption{Fixed-hardware generation scaling for PPO and GRPO. Blue bars show training responses collected per optimization boundary and orange bars show boundary-collection time, each normalized to the smallest batch. Boxed labels report the corresponding generation-stage response throughput.}
    \label{fig:fixed-hardware-generation-scaling}
\end{figure}

\paragraph{PPO: response-collection throughput improves by $2.29\times$.}
Figure~\ref{fig:fixed-hardware-generation-scaling}(a) reports generation-stage measurements from asynchronous partial-rollout implementations. In the Hunyuan A3B PPO runs, increasing $B_{\mathrm{train}}$ from 256 to 1024 multiplies responses per optimization boundary by $4\times$, while boundary-collection time grows only from 39 to 68 seconds ($68/39=1.74\times$). The resulting response-collection throughput improves by $4/(68/39)=2.29\times$. This suggests that rollout collection is underfilled at the smaller PPO batch, so increasing the response batch can improve generation throughput even on fixed hardware.

\paragraph{GRPO: response-collection throughput improves by up to $1.36\times$.}
Figure~\ref{fig:fixed-hardware-generation-scaling}(b) shows the same sublinear collection-time scaling in the Qwen3-30B-A3B GRPO runs. Over the common 40,960--286,720 training-response window at fixed $G=8$, increasing $P$ from 128 to 256 doubles the responses per optimization boundary but increases boundary-collection time by only $1.53\times$, yielding a $1.31\times$ response-throughput gain. Increasing $P$ to 512 raises responses by $4\times$ while time grows by $2.93\times$, yielding a $1.36\times$ gain. The small additional improvement from $P=256$ to $P=512$ suggests diminishing generation-throughput returns over this range.

\paragraph{Actor-update time is approximately batch-proportional.}
The same GRPO measurements show a different scaling pattern for optimization. At fixed $G=8$, doubling $P$ from 128 to 256 increases mean actor-update time from 101.3 to 208.6 seconds ($2.06\times$) under the scaled learning rate; the fixed-LR control gives a similar $1.98\times$ increase. Thus, over the measured range, actor-update time is close to linear in the training batch, in contrast to the sublinear growth of rollout-collection time.

\subsection{Sublinear Generation Cost and the Critical Generation Batch}
\label{sec:sublinear-generation-cost}
\label{sec:decode-roofline}

The empirical asymmetry in Section~\ref{sec:generation-evidence} comes from how fixed hardware executes the two stages. Autoregressive generation advances each sequence by one token at a time; token $u+1$ cannot be computed before token $u$, so a single sequence exposes little parallel work. At low active concurrency, each layer applies large weight matrices to only a few token vectors. Decode therefore has low arithmetic intensity and is often memory-bandwidth-bound. Increasing the number of concurrent sequences widens these batched operations and amortizes each weight transfer over more generated tokens, raising hardware utilization and throughput \citep{williams2009roofline,pope2023efficiently,zhong2024distserve}.

Training exposes substantially more parallelism. Once the responses are available, the forward and backward passes process many token positions across many sequences simultaneously. The resulting large matrix--matrix operations have higher arithmetic intensity and typically keep the arithmetic units much better utilized. On fixed hardware, increasing the training batch therefore adds approximately proportional FLOPs rather than unlocking the same unused parallelism. Consistent with this distinction, Section~\ref{sec:generation-evidence} observes sublinear rollout-collection time but nearly batch-proportional actor-update time.

We use $B_{\mathrm{gen}}$ for rollout responses available to generation and $B_{\mathrm{train}}$ for retained responses consumed by an optimizer update. We therefore model training with a batch-proportional term and generation with an amortized term:
\begin{equation}
    t_{\mathrm{gen}}(B_{\mathrm{gen}})\approx d_0+d_1B_{\mathrm{gen}},
    \qquad
    t_{\mathrm{train}}(B_{\mathrm{train}})\approx \gamma B_{\mathrm{train}}.
    \label{eq:simple-generation-training-time}
\end{equation}
Here $d_0$ is the batch-independent generation latency, $d_1$ is the batch-proportional generation-time coefficient, and $\gamma$ is the batch-proportional training-time coefficient; $d_1$ and $\gamma$ have units of time per response under the chosen batch unit. The intercept captures work amortized across concurrent responses. Generation time can therefore grow sublinearly, whereas training time scales roughly with batch.

These amortization gains eventually saturate as batch-dependent arithmetic, KV-cache traffic, response-length imbalance, and memory capacity become limiting \citep{yu2022orca,pope2023efficiently,kwon2023pagedattention}. We call the smallest logical rollout batch that captures most of the attainable generation throughput the \emph{critical generation batch}; Appendix~\ref{sec:analytical-systems-details} derives this saturation with a decode-step Roofline model and relates the logical rollout batch to the active batch seen by decode kernels.

\begin{defn}[Critical generation batch]
\label{def:critical-generation-batch}
For a fixed model, decoding engine, length distribution, and hardware allocation, let $q_{\mathrm{gen}}(B)$ be aggregate generation throughput at batch $B$, and let $q_{\mathrm{gen}}^{\max}$ be its plateau. For a small tolerance $\delta>0$, the critical generation batch $B_{\mathrm{gen}}^{\mathrm{crit}}$ is the smallest $B$ satisfying
\begin{equation*}
    q_{\mathrm{gen}}(B)\geq(1-\delta)q_{\mathrm{gen}}^{\max}.
\end{equation*}
\end{defn}

\paragraph{Practical implication.}
Definition~\ref{def:critical-generation-batch} marks the systems knee beyond which most attainable generation throughput has already been realized; it is not by itself the wall-clock-optimal batch. Unlike $B_{\mathrm{train}}^{\mathrm{crit}}$, $B_{\mathrm{gen}}^{\mathrm{crit}}$ depends on the model, decoding engine, response-length distribution, and hardware allocation, and must be profiled for the target deployment. In practice, sweep $B_{\mathrm{gen}}$ or decode concurrency, verify stable throughput and KV-cache feasibility, and stop once additional batching yields little throughput gain. The selected configuration lowers time-to-target only if its end-to-end throughput gain exceeds the samples-to-target penalty. Section~\ref{sec:final-guidance} combines these generation- and training-side constraints.

\begin{tcolorbox}[
  title={Summary and takeaway},
  fonttitle=\bfseries,
  colback=RoyalBlue!4,
  colframe=RoyalBlue!55!black,
  boxrule=0.6pt,
  arc=1.5mm,
  left=2mm,
  right=2mm,
  top=1mm,
  bottom=1mm
]
\begin{itemize}
  \setlength{\topsep}{0pt}
  \setlength{\itemsep}{0.15em}
  \setlength{\parsep}{0pt}
  \item On fixed hardware, increasing the response batch improves generation throughput by up to $2.29\times$ in PPO and $1.36\times$ in GRPO.
  \item Generation and training scale asymmetrically: larger active batches amortize model-weight traffic during decoding, whereas training work grows approximately with processed tokens.
  \item Generation time therefore often grows sublinearly with the response batch, until another systems bottleneck causes throughput to plateau.
\end{itemize}
\end{tcolorbox}

\section{Operating Regimes and Practical Guidance}
\label{sec:final-guidance}

The preceding sections separately measured learning per sample and systems throughput. We now combine them through $T=N^\star/q$: a larger batch reduces time-to-target only when its throughput gain exceeds its samples-to-target penalty $r_N$. In principle, this tradeoff can take one of two forms, depending on whether the generation or training critical batch is reached first.

\begin{figure}[htbp]
    \centering
    \includegraphics[width=\textwidth]{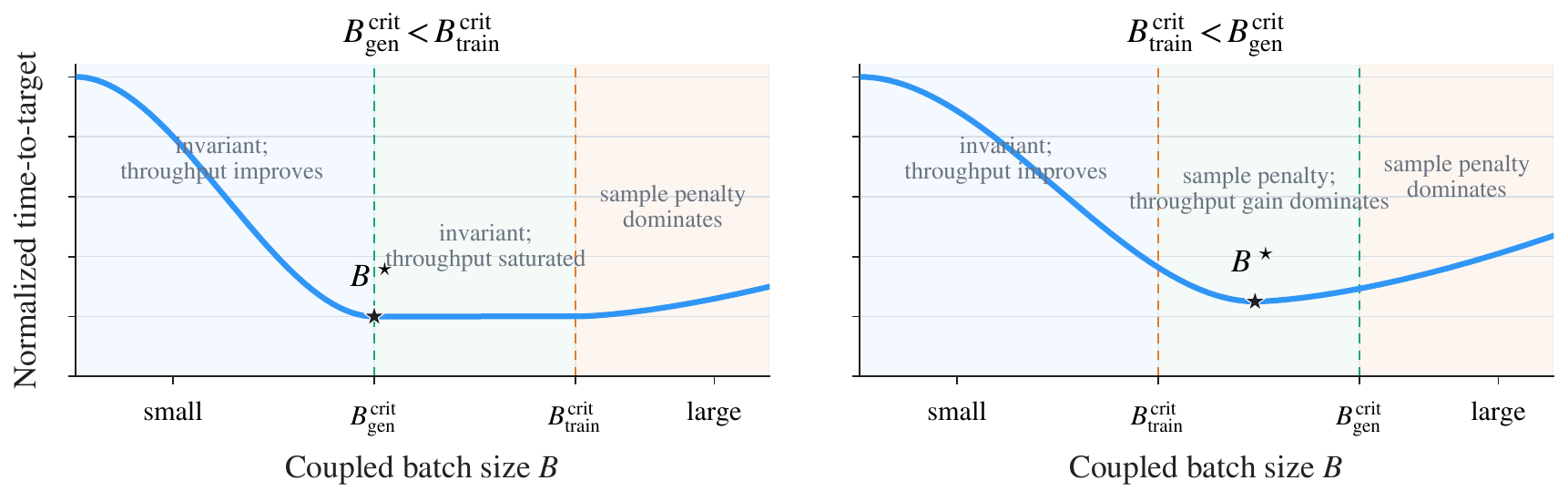}
    \caption{Schematic normalized time-to-target for a synchronously coupled batch $B$. The panels show the two possible orderings of $B_{\mathrm{train}}^{\mathrm{crit}}$ and $B_{\mathrm{gen}}^{\mathrm{crit}}$; $B^\star$ marks the wall-clock optimum. The curves are schematic rather than fitted to our experiments.}
    \label{fig:time-to-target-scenarios}
\end{figure}

\paragraph{Two orderings of the critical batches.}
Which ordering arises depends on the hardware, model size, and dataset size. When one coupled batch $B$ scales both generation and training, the two orderings produce two practical regimes. Figure~\ref{fig:time-to-target-scenarios} illustrates the resulting $T(B)=N^\star(B)/q(B)$; neither critical batch alone determines the minimizing batch $B^\star$.

\begin{itemize}
    \setlength{\topsep}{0.25em}
    \setlength{\itemsep}{0.25em}
    \setlength{\parsep}{0pt}
    \item \textbf{Left: generation knee first.} Throughput plateaus while $r_N\approx1$, creating a broad near-optimal region. Further scaling provides little systems benefit and eventually raises time-to-target as the sample penalty becomes material.
    \item \textbf{Right: training knee first.} The sample penalty begins while generation throughput is still improving. When the throughput gain is larger, $B^\star$ can lie beyond $B_{\mathrm{train}}^{\mathrm{crit}}$ but before $B_{\mathrm{gen}}^{\mathrm{crit}}$.
\end{itemize}
\FloatBarrier

\paragraph{Measured GRPO operating point.}
Table~\ref{tab:grpo-target-accounting} applies this accounting to the fixed-$G=8$ GRPO sweep. It reports both sides of the tradeoff for every configuration: $r_N$ measures the samples-to-target penalty, while $q/q_{128}$ measures the realized throughput gain relative to $P=128$.

\begin{figure}[!t]
    \centering
    \includegraphics[width=0.70\textwidth]{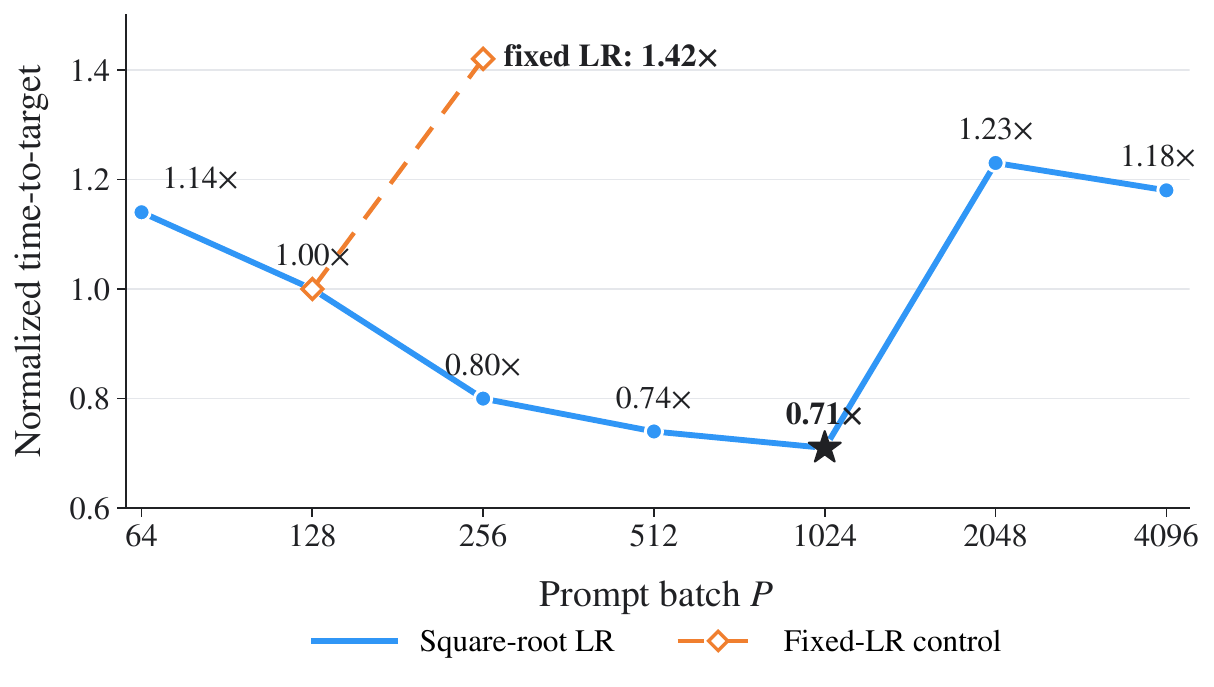}
    \caption{Measured GRPO time-to-target sweep at fixed $G=8$ and $J^\star=77\%$, using the accounting in Table~\ref{tab:grpo-target-accounting}. Blue shows the square-root-LR configurations, orange shows the fixed-LR control, and the star marks the minimum measured time-to-target. All times are normalized to the $P=128$ reference.}
    \label{fig:grpo-time-to-target-sweep}
\end{figure}

\begin{table}[htbp]
    \centering
    \caption{GRPO time-to-target accounting at fixed $G=8$ and $J^\star=77\%$. $K^\star$ is the evaluated optimizer update used for accounting, and $J(K^\star)$ is its validation value. $N^\star_{\mathrm{train}}$ is reported in thousands of retained training responses, $q=N^\star_{\mathrm{train}}/T$, and all ratios use $P=128$ as the reference. The final row is the fixed-LR $P=256$ control; all other rows use square-root LR scaling.}
    \label{tab:grpo-target-accounting}
    \begingroup
    \begin{tabular*}{\columnwidth}{@{\extracolsep{\fill}}rrrrrrrr@{}}
        \toprule
        $P$ & $\eta/10^{-6}$ & $J(K^\star)$ & $K^\star$ & $N^\star_{\mathrm{train}}$ (K) & $T$ (h) & $r_N$ & $q/q_{128}$ \\
        \midrule
        64   & $1/\sqrt{2}$ & $78.80$ & 200 & 102.40 & 13.56 & 0.83 & 0.73 \\
        128  & $1$          & $77.50$ & 120 & 122.88 & 11.90 & 1.00 & 1.00 \\
        256  & $\sqrt{2}$   & $76.72$ & 60  & 122.88 & 9.54 & 1.00 & 1.25 \\
        512  & $2$          & $78.59$ & 30  & 122.88 & 8.77  & 1.00 & 1.36 \\
        1024 & $2\sqrt{2}$  & $77.19$ & 15  & 122.88 & 8.42  & 1.00 & 1.41 \\
        2048 & $4$          & $79.58$ & 12  & 196.61 & 14.68 & 1.60 & 1.30 \\
        4096 & $4\sqrt{2}$  & $78.02$ & 6 & 196.61 & 14.00 & 1.60 & 1.36 \\
        \midrule
        256  & $1$ fixed    & $77.40$ & 100 & 204.80 & 16.93 & 1.67 & 1.17 \\
        \bottomrule
    \end{tabular*}
    \endgroup
\end{table}
Figure~\ref{fig:grpo-time-to-target-sweep} visualizes the same accounting. The measured sweep follows the generation-knee-first pattern in Figure~\ref{fig:time-to-target-scenarios}: normalized time-to-target falls to $0.71$ at $P=1024$ while $r_N$ remains near one, then rises to $1.23$ and $1.18$ at $P=2048$ and $P=4096$ once $r_N$ reaches $1.60$. The fixed-LR $P=256$ control takes $1.42\times$ as long as the reference because its $1.17\times$ throughput gain does not offset its $1.67\times$ sample penalty.

\begin{principle}[Align, then accelerate]
\label{princ:align-then-accelerate}
First retune batch-dependent hyperparameters to preserve learning per sample. Then increase batching only while its throughput gain exceeds its samples-to-target penalty.
\end{principle}

\paragraph{Two-stage tuning procedure.}
Algorithm~\ref{alg:joint-batch-tuning} provides a practical procedure for applying Principle~\ref{princ:align-then-accelerate}. It first establishes training-side alignment and then searches for systems-side throughput gains.

\begin{algorithm}[H]
\caption{Two-stage batch tuning}
\label{alg:joint-batch-tuning}
\begin{algorithmic}[1]
\Require Reference $(B_0,\eta_0)$, candidate training batches $\mathcal B$, and target $J^\star$
\For{$B_{\mathrm{train}}\in\mathcal B$}
    \State For Adam-type optimizers, initialize $\eta\gets\eta_0\sqrt{B_{\mathrm{train}}/B_0}$; retune as needed.
    \State Run a sample-matched learning curve and estimate $r_N$ at $J^\star$.
    \State Increase $B_{\mathrm{gen}}$ or decode concurrency until throughput plateaus or a feasibility limit is reached.
    \State Measure stable end-to-end throughput $q$ using the same sample counter.
\EndFor
\State \Return the feasible configuration maximizing $q/r_N$.
\end{algorithmic}
\end{algorithm}

\begin{samepage}
For each candidate training batch, the first stage initializes and retunes batch-dependent hyperparameters, then estimates $r_N$ from a sample-matched learning curve. We use the square-root learning-rate initialization for Adam-type optimizers; other optimizers require their corresponding batch-scaling rule. The second stage increases generation concurrency and measures stable end-to-end throughput $q$. The procedure returns the feasible configuration maximizing $q/r_N$.
\end{samepage}

\section{Related Work}
\label{sec:related-work}

The closest literature addresses three quantities that are often conflated: the training batch used to estimate an update, the number of rollouts used to explore each prompt, and the active concurrency used to execute generation. We keep these quantities separate and reconnect them only through time-to-target.

\subsection{Training-Batch Scaling and Invariance}

Classical critical-batch work is motivated by data-parallel training when examples are already available: increasing the batch can reduce the number of sequential optimizer updates, but this benefit exhibits diminishing returns. \citet{mccandlish2018empirical} relate the transition to the gradient noise scale, while \citet{shallue2019measuring} show empirically that useful parallelism is workload-dependent and that batch comparisons require fair hyperparameter tuning. The critical batch can also change over the course of language-model training \citep{merrill2025critical}.

\citet{hilton2022batch} define policy optimization as batch-size invariant when hyperparameter retuning preserves behavior as a function of processed samples. They show that standard PPO is not automatically invariant because its old policy serves both behavior-correction and proximal roles. Their PPO-EWMA and PPG-EWMA constructions decouple these roles, use square-root learning-rate scaling for Adam, and require a single policy epoch for their invariance construction; they separately show that the decoupled objective better tolerates artificially stale data. The SDE analysis of \citet{malladi2022sde} provides a complementary account of the Adam scaling rule.

Relative to this work, our contribution is an empirical transfer test in a different combined regime rather than a new invariance construction. We study modern LLM RL using single-pass policy-gradient methods: both GRPO and PPO perform one global update per rollout batch, without minibatch steps or rollout reuse, and both workloads run with asynchronous partial rollouts and streaming generation. Without introducing the EWMA proximal-policy mechanism of \citet{hilton2022batch}, we test how far Adam learning-rate retuning alone aligns their sample-indexed learning curves. The resulting evidence supports bounded approximate invariance in this LLM RL setting and connects it to measured systems throughput and time-to-target.

\subsection{Rollout Scaling and Generation Efficiency}

Our framework is a fixed-hardware extension of classical critical-batch analysis \citep{mccandlish2018empirical,shallue2019measuring}: it compares the statistical cost of a larger batch, measured by samples-to-target, with its systems benefit, measured by end-to-end throughput. \citet{piche2025pipelinerl} use a similar effectiveness--throughput decomposition to study pipeline scheduling and data staleness, whereas we use it to characterize batch-size selection under fixed hardware.

The work by \citet{hu2025brorl} is the closest to our outcome-filtering setting. BroRL increases the number of rollouts per prompt from 16 to 512, making a group less likely to be rejected as uniformly correct or uniformly incorrect. It reports both higher dynamic-sampling acceptance and higher generation throughput, and interprets the algorithmic benefit more broadly as improved exploration. The key conceptual difference is that we formally study batch-size invariance. In our framework, both the prompt batch $P$ and group size $G$ are batch dimensions: after appropriate retuning, changing either can approximately preserve sample-indexed learning. BroRL instead emphasizes a mechanism specific to dynamic sampling, where a larger rollout count per prompt reduces all-correct and all-incorrect groups and thereby increases the fraction of generated responses retained for training. Related methods, including Knapsack RL, AR3PO, and VIP, likewise improve sampling efficiency by allocating prompt-dependent rollout budgets rather than increasing $G$ uniformly \citep{li2026knapsack,zhang2025ar3po,nguyen2026vip}. These methods improve the learning value obtained from a fixed rollout budget rather than establishing invariance of learning per retained response across batch configurations.

The IsoCompute Playbook optimizes how a fixed generated-rollout budget is allocated across problems per batch, rollouts per problem, and sequential updates \citep{cheng2026isocompute}. Its preference for larger rollout groups partly reflects fewer zero-signal groups on hard problems, but also solution sharpening and reduced cross-problem interference; unlike our retained-response analysis, it does not separate sampling yield from learning per retained response.

On the systems side, the Roofline model explains why increasing active decode concurrency can amortize weight traffic \citep{williams2009roofline}, while KV-cache management constrains feasible concurrency \citep{kwon2023pagedattention}. Frameworks that overlap or reschedule rollout and training stages change the attainable end-to-end throughput \citep{sheng2025hybridflow,fu2025areal,piche2025pipelinerl}; they do not change the criterion that the throughput gain must exceed the sample penalty.

\section{Conclusion}
\label{sec:conclusion}

This paper studies when larger batches help scale LLM reinforcement learning by separating learning per sample from systems throughput. After batch-dependent hyperparameters are retuned, batch-size invariance makes different physical batches comparable through approximately aligned sample-indexed learning curves over a bounded range. Our GRPO and PPO experiments support this property, while the fixed-learning-rate control and the largest-batch runs show that it is not automatic or unbounded. Generation--training asymmetry provides the complementary systems opportunity: increasing rollout concurrency can improve response throughput without proportionally increasing generation time, and our tuned GRPO configurations translate this gain into lower time-to-target. The practical takeaway is therefore simple: retune the learning rate and other batch-dependent hyperparameters whenever the training batch changes; once the recipe is stable, increase the batch only while its end-to-end throughput gain exceeds its samples-to-target penalty.

Because our experiments retune only the learning rate, future work should derive scaling rules for other batch-dependent hyperparameters and pursue tighter batch-size invariance as a basis for algorithm design, comparison, and transfer. It should also measure how the training and generation critical batches shift across models, algorithms, hardware, and execution modes, and clarify the coupled actor--critic batch scales in PPO.

\section*{Author Contributions}

Ziniu Li led the project and developed the central connection between batch-size invariance and generation--training computational asymmetry. Jinbo Wang, Guanhua Huang, and Alex Chen contributed to the conceptual discussions. In particular, Jinbo Wang noted the expected batch scaling of the diagnostic actor gradient norm and the differing behavior of actor and value-model gradients in PPO. Feiyuan Zhang and Pengbo Li analyzed and interpreted the computational asymmetry between rollout generation and training. Alex Chen supervised the project.

\bibliography{main}
\bibliographystyle{colm2024_conference}

\appendix

\section{Analytical Details for Generation--Training Asymmetry}
\label{sec:analytical-systems-details}

This appendix collects the analytical details omitted from Section~\ref{sec:decode-roofline}. For a synchronous comparison with a fixed response-length distribution, set $B_{\mathrm{gen}}=B_{\mathrm{train}}=B$ and consider a fixed budget of $N$ responses. Under Equation~\eqref{eq:simple-generation-training-time}, total generation and training time are approximately
\begin{align*}
    T_{\mathrm{gen}}(N;B)
    &\approx \frac{N}{B}\,t_{\mathrm{gen}}(B)
    \approx N\left(\frac{d_0}{B}+d_1\right),
    \\
    T_{\mathrm{train}}(N;B)
    &\approx \frac{N}{B}t_{\mathrm{train}}(B)
    \approx N\gamma.
\end{align*}
The repeated weight-streaming term decreases as $d_0/B$, while the dominant training arithmetic remains approximately constant over the fixed response budget.

The aggregate generation batch in the main text is not necessarily the batch dimension seen by an individual decode kernel. At decode step $u$, let $B_{\mathrm{act}}(u)\leq B_{\mathrm{gen}}$ be the number of unfinished sequences that contribute a new token. The distribution of $B_{\mathrm{act}}(u)$ over a rollout boundary is determined by admission, scheduling, and the response-length distribution. Increasing $B_{\mathrm{gen}}$ improves generation throughput only insofar as it raises this active concurrency.

A decode-step Roofline calculation makes the generation-side amortization explicit. At step $u$, the system produces one new token for each of the $B_{\mathrm{act}}(u)$ active sequences. The same weights are applied to these tokens in a batched operation, so weight traffic is largely independent of $B_{\mathrm{act}}(u)$, whereas arithmetic and KV-cache traffic grow with it \citep{pope2023efficiently,zhong2024distserve}. Let $M_W$ be the weight bytes transferred at a decode step, $m_u$ the additional per-sequence memory traffic at step $u$, including KV-cache reads and writes, $f_{\mathrm{tok}}$ the FLOPs per generated token, $W_{\mathrm{HBM}}$ the sustained memory bandwidth, and $F_{\mathrm{eff}}$ the sustained arithmetic throughput. The Roofline model gives \citep{williams2009roofline}
\begin{equation}
    t_{\mathrm{dec}}(u)
    \gtrsim
    \max\!\left\{
        \frac{M_W+B_{\mathrm{act}}(u)m_u}{W_{\mathrm{HBM}}},
        \frac{B_{\mathrm{act}}(u)f_{\mathrm{tok}}}{F_{\mathrm{eff}}}
    \right\},
    \qquad
    q_{\mathrm{tok}}(u)=\frac{B_{\mathrm{act}}(u)}{t_{\mathrm{dec}}(u)}.
    \label{eq:decode-roofline-time}
\end{equation}
Here $t_{\mathrm{dec}}(u)$ is decode-step latency and $q_{\mathrm{tok}}(u)$ is the corresponding token throughput. At small $B_{\mathrm{act}}(u)$, weight transfer dominates, so $t_{\mathrm{dec}}(u)\approx M_W/W_{\mathrm{HBM}}$ changes little as active concurrency increases, while $q_{\mathrm{tok}}(u)$ grows nearly linearly. At larger active batches, the batch-dependent compute or KV-cache term eventually dominates. Step time then grows with active concurrency, and throughput approaches a plateau instead of continuing to scale linearly.

Equation~\eqref{eq:decode-roofline-time} also gives a hardware lower bound for each decode step. The corresponding decode arithmetic intensity and ridge point are
\begin{equation*}
    \mathrm{AI}_{\mathrm{dec}}(u)
    =\frac{B_{\mathrm{act}}(u)f_{\mathrm{tok}}}
    {M_W+B_{\mathrm{act}}(u)m_u},
    \qquad
    \rho=\frac{F_{\mathrm{eff}}}{W_{\mathrm{HBM}}}.
\end{equation*}
Decode is bandwidth-bound when $\mathrm{AI}_{\mathrm{dec}}(u)<\rho$. Increasing the active batch amortizes $M_W$ and raises arithmetic intensity until the ridge point or another limit is reached.

\begin{figure}[htbp]
    \vspace{-0.5em}
    \centering
    \includegraphics[width=0.95\textwidth]{%
        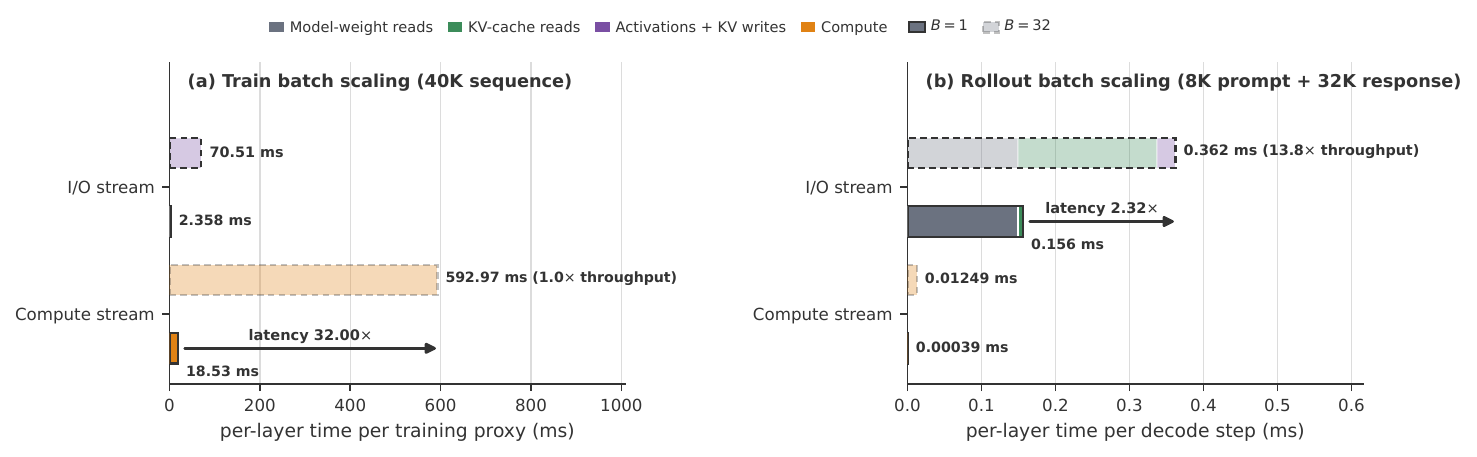}
    \caption{Analytical per-layer stream decomposition under batch scaling for (a) a training proxy and (b) rollout in a representative fixed-hardware setting. Solid and light dashed bars denote $B=1$ and $B=32$, respectively. Assuming ideal overlap between I/O and compute, latency is modeled as the maximum of the two; parenthesized values report ideal aggregate throughput relative to $B=1$.}
    \label{fig:dense-stream-batch-scaling}
    \vspace{-0.5em}
\end{figure}

Figure~\ref{fig:dense-stream-batch-scaling} instantiates the Roofline model for a 110B-parameter dense transformer with tensor-parallel degree $\mathrm{TP}=4$, using parameters representative of a high-bandwidth accelerator. Panel~(a) uses a forward-pass GEMM proxy with 40K tokens per sequence, whereas panel~(b) averages decode steps over an 8K-token prompt and a 32K-token response. Accordingly, the plotted symbol $B$ denotes the local training batch in panel~(a) and the active decode batch in panel~(b). The common sweep from 1 to 32 provides a controlled comparison; it does not imply that $B_{\mathrm{train}}=B_{\mathrm{act}}$ in an asynchronous deployment. The resulting values are analytical per-layer lower bounds. They are not end-to-end measurements and do not estimate the critical batch.

Increasing the local training batch from 1 to 32 raises the compute-bound training-proxy latency by $32\times$, leaving ideal aggregate training throughput unchanged. By contrast, increasing the active decode batch by $32\times$ raises the I/O-bound decode latency by only $2.32\times$, because batch-independent model-weight traffic is amortized across the batch. This yields a $13.8\times$ ideal aggregate decode throughput gain. The figure illustrates this mechanism; the absolute times remain configuration-dependent.

Operationally, $B_{\mathrm{gen}}^{\mathrm{crit}}$ is the logical batch whose active-concurrency distribution brings aggregate throughput near its plateau, rather than a universal kernel batch. KV-cache capacity imposes a separate feasibility limit. For context length $S$, number of transformer layers $L_{\mathrm{layer}}$, number of KV heads assigned to each device after tensor-parallel sharding $H_{\mathrm{KV}}$, head dimension $d_h$, and bytes per cached scalar $s_{\mathrm{KV}}$, the per-sequence KV-cache footprint and an optimistic memory-feasible active batch are
\begin{equation*}
    m_{\mathrm{KV}}(S)
    =2L_{\mathrm{layer}}H_{\mathrm{KV}}d_hS s_{\mathrm{KV}},
    \qquad
    B_{\mathrm{KV}}
    =\left\lfloor
    \frac{\alpha C_{\mathrm{mem}}-M_{\mathrm{weights}}-M_{\mathrm{runtime}}}
    {m_{\mathrm{KV}}(S)}
    \right\rfloor,
\end{equation*}
where $C_{\mathrm{mem}}$ is device memory capacity, $M_{\mathrm{weights}}$ is the resident model-weight footprint, $\alpha\in(0,1)$ is the fraction available to the serving engine, and $M_{\mathrm{runtime}}$ reserves memory for activations, workspaces, fragmentation, and runtime state. Under tensor parallelism, all terms are per-device after sharding. Here $M_{\mathrm{weights}}$ is a resident-capacity term, whereas $M_W$ in Equation~\eqref{eq:decode-roofline-time} denotes weight traffic during a decode step. Paging reduces fragmentation but not the linear growth with context length \citep{kwon2023pagedattention}. A feasible logical generation batch must induce $B_{\mathrm{act}}(u)\leq B_{\mathrm{KV}}$ throughout decoding, subject to runtime admission and stability constraints; this memory limit may prevent aggregate generation throughput from reaching its plateau.

\end{document}